\pdfoutput=1

\documentclass[11pt]{article}

\usepackage[final]{acl}

\usepackage{times}
\usepackage{latexsym}

\usepackage[T1]{fontenc}

\usepackage[utf8]{inputenc}

\usepackage{microtype}

\usepackage{inconsolata}

\usepackage{graphicx}
\usepackage{algorithm}
\usepackage{threeparttable}
\usepackage{multirow,booktabs,amssymb,amsmath}

\usepackage{hyperref}
\usepackage{colortbl}        
\usepackage{array}

\usepackage{algpseudocode}
\usepackage{amsfonts}

\DeclareMathOperator*{\argmax}{argmax}

\title{Beyond Stochastic Exploration: What Makes Training Data Valuable for Agentic Search}

\author{
\textbf{Chuzhan Hao, Wenfeng Feng, Guochao Jiang, Guofeng Quan,}\\
\textbf{Guohua Liu, and Yuewei Zhang\protect  \thanks{Corresponding author.}}\\
Alibaba Cloud Computing\\
\texttt{\{haochuzhan.hcz, liyou.zyw\}@alibaba-inc.com}
}

\begin{document}
\maketitle

\begin{abstract}
Reinforcement learning (RL) has become an effective approach for advancing the reasoning capabilities of large language models (LLMs) through the strategic integration of external search engines. However, current RL-based search agents often rely on a process of stochastic exploration guided by carefully crafted outcome rewards, leading to inefficient reasoning trajectories and unstable training.
To address these issues, we propose a novel framework, Hierarchical Experience (HiExp), to enhance the performance and training stability of search agents.
Specifically, we extract empirical knowledge through contrastive analysis and a multi-level clustering mechanism, transforming raw reasoning trajectories into hierarchical experience knowledge. By leveraging experience-aligned training, we effectively regularize stochastic exploration, evolving it into a strategic and experience-driven search process.
Extensive evaluations on multiple complex agentic search and mathematical reasoning benchmarks demonstrate that our approach not only achieves substantial performance gains but also exhibits strong cross-task and cross-algorithm generalization.

\end{abstract}

\section{Introduction}
Large language models have demonstrated remarkable capabilities in task planning and agentic reasoning, with reinforcement learning significantly improving their performance on complex reasoning tasks~\cite{shao2024deepseekmath,guo2025deepseek,yang2025qwen3}. However, reliance on static parametric knowledge presents notable limitations, often leading to hallucinations and inefficient reasoning~\cite{yao2025reasoning,kalai2025hallucinate}. To tackle these challenges, it is crucial to explore how to efficiently access diverse external information to support LLMs in achieving deliberate and well-substantiated reasoning. Therefore, a novel search paradigm termed \textit{Agentic Deep Research Systems} has gradually become an important research task~\cite{li2025searcho1,jin2025searchr1,zhang2025websearch}.

\begin{figure}[t]
\centering
\includegraphics[width=0.49\textwidth]{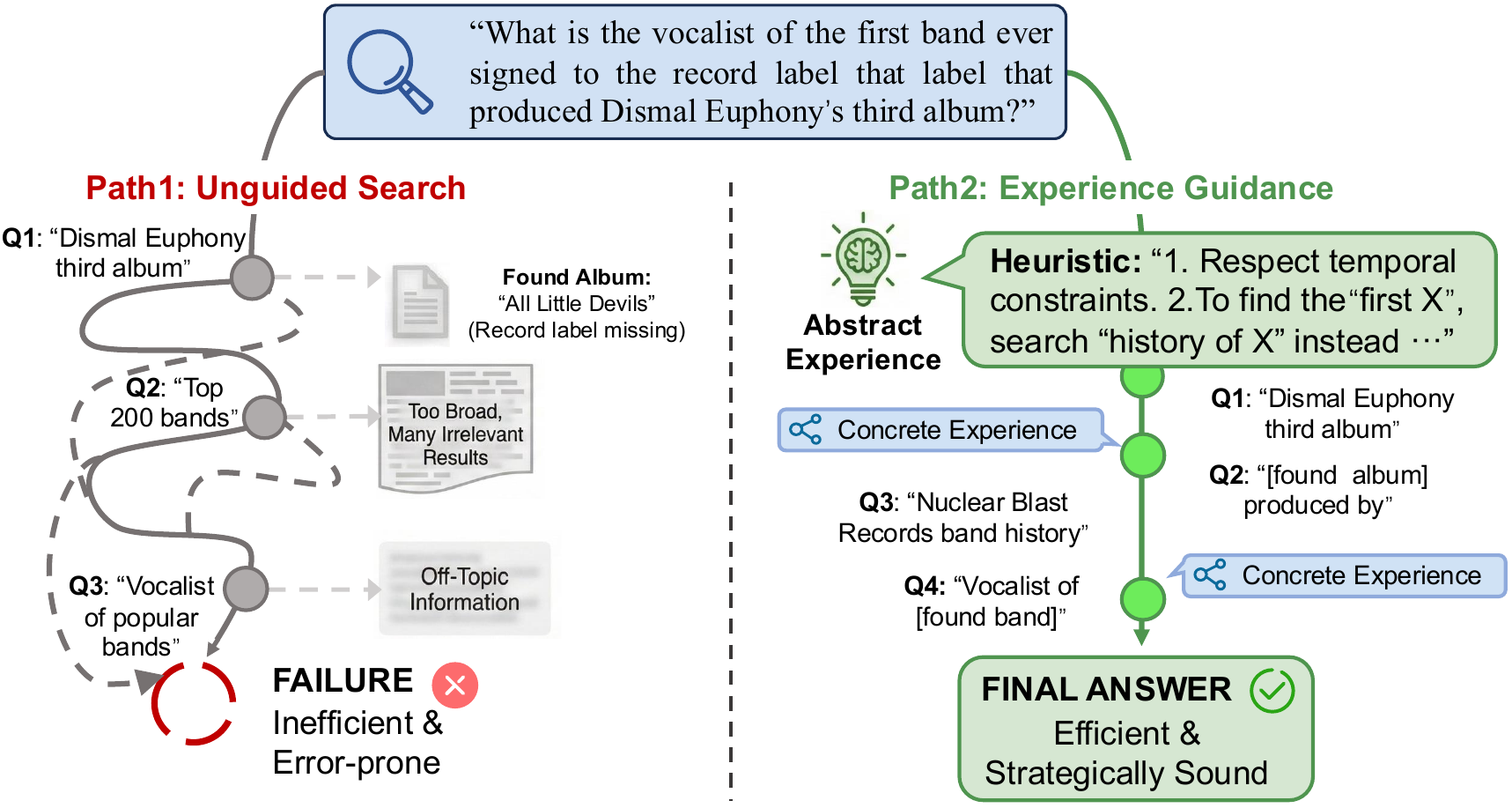} 
\caption{Comparison between stochastic exploration and experience-guided exploration. Experience-driven guidance facilitates more efficient reasoning trajectories, endowing LLMs with superior problem-solving capabilities for complex tasks.}
\label{placeholder}
\end{figure}

Previous research has utilized Chain-of-Thought (CoT)~\cite{wei2022chainofthought} prompting to decompose complex problems into sequential sub-tasks, subsequently leveraging external information dynamically to bridge knowledge gaps and tackle intricate reasoning tasks~\cite{trivedi2022ircot,yue2024iterdrag,feng2025airrag}. \citet{li2025searcho1} integrates agentic search into the reasoning process, enabling dynamic retrieval to address informational uncertainty or incompleteness. Recently, reinforcement learning has achieved remarkable success in mathematical reasoning and decision-making scenarios~\cite{guo2025deepseek}. \citet{jin2025searchr1,feng2025retool} also utilize RL through environmental interactions to significantly enhance the capability of small language models (SLMs) in addressing intricate multi-hop and mathematical reasoning challenges. These training-based approaches integrate autonomous tool invocation into LLMs,
facilitating dynamic environmental interaction~\cite{zheng2025deepresearcher,chen2025research}. Due to their superior agentic abilities and strong generalization, RL-based agentic reasoning approaches are increasingly emerging as a significant trend in deep research~\cite{zhang2025websearch}. 

Existing RL-based search agents rely primarily on stochastic exploration guided by carefully crafted outcome rewards. However, these methods often struggle to execute global strategic planning and explore efficient reasoning trajectories, particularly when utilizing small language models for complex complex tasks, as shown in Figure~\ref{placeholder}. Furthermore, in multi-turn interaction scenarios, the inherent difficulty of providing consistent reward signals leads to significant instability during end-to-end RL training.

To address these limitations, we introduce the HiExp framework, which regularizes the exploration process of search agents with hierarchical experiences. By transforming stochastic exploration into a strategic, experience-aligned search, we significantly stabilize the reward signals and facilitate the discovery of optimal reasoning paths.
Specifically, we extract empirical knowledge by performing contrastive analysis on pre-sampled rollouts, identifying the critical factors that differentiate successful reasoning paths from failures. We then employ a multi-level clustering strategy to abstract these instance-specific insights into high-dimensional reasoning strategies. 
These hierarchical experiences significantly bolster LLMs' performance across diverse task scenarios during the inference phase.
Furthermore, throughout the critic-free RL training process, these systemic experiences are dynamically aligned with the rollout stage. This alignment effectively transforms conventional stochastic exploration into a strategic, experience-driven search, enhancing the effectiveness and stability of the optimization process.
In summary, our main contributions are as follows:

\begin{itemize}
    \item We introduce an endogenous scheme for hierarchical experience (HiExp) construction by leveraging self-reflection and agglomerative clustering over internal reasoning trajectories. This method facilitates the autonomous synthesis of meta-knowledge without the need for additional external factual information.

    \item Our proposed HiExp not only improves LLMs' performance in various tasks during the inference phase but also dynamically aligns with the rollout stage of RL training. This alignment transforms conventional stochastic exploration into a strategic, experience-driven search, enhancing the effectiveness and stability of policy optimization.

    \item Extensive evaluations demonstrate that HiExp consistently yields substantial performance gains over RL-based search agents. Furthermore, our approach exhibits robust generalization capabilities across various task domains and RL algorithms.

\end{itemize}

\section{Related Work}
\subsection{Retrieval-Augmented Generation}
Early retrieval-augmented generation (RAG) approaches employ various strategies such as branching, iteration, and adaptive retrieval to solve complex tasks. These methods rely on manually crafted workflows to guide LLMs in interacting with external knowledge sources. IRCoT~\cite{trivedi2022ircot} leverages CoT to steer the retrieval process, refining CoT with the retrieved information. \citet{press2022selfask,asai2023selfrag,yue2024iterdrag} refine intermediate queries to acquire valuable knowledge through multi-turn iterations. AirRAG~\cite{feng2025airrag} applies Monte Carlo Tree Search to dynamically explore the reasoning paths. However, these approaches are limited to manually designed prompts and workflows, failing to fully unleash the inherent reasoning potential of LLMs.

\subsection{Autonomous Search Agents}
As the reasoning and decision-making capabilities of the foundation models continue to improve, Search-o1~\cite{li2025searcho1} significantly improves model performance in complex scenarios by designing an agentic search workflow, providing superior flexibility and generalization. DeepSeek-R1~\cite{guo2025deepseek} also demonstrates that outcome-based RL can significantly enhance the autonomous reasoning and decision-making capabilities of models. Therefore, RL has been applied to various complex reasoning tasks and agent-based scenarios. Complex multi-hop question answering represents a typical integrated application scenario that heavily relies on model-driven planning and reasoning. \citet{chen2025research,jin2025searchr1,feng2025pvpo} have successfully applied end-to-end RL to complex agentic search scenarios, further advancing the development of agentic deep research systems. These methods autonomously select retrieval tools during the reasoning process to interact with external environments.
DeepResearcher~\cite{zheng2025deepresearcher} scales RL in real-world environments by incorporating authentic web search interactions. s3~\cite{jiang2025s3} decouples the searcher from the generator and trains the searcher with fewer samples. EvolveSearch~\cite{zhang2025evolvesearch} further explores the self-evolution process of search agents. StepSearch~\cite{wang2025stepsearch} introduces fine-grained reward signals to steer strategic query planning and improve retrieval quality in complex search environments.

\begin{figure*}[t]
\centering
\includegraphics[width=1.00\textwidth]{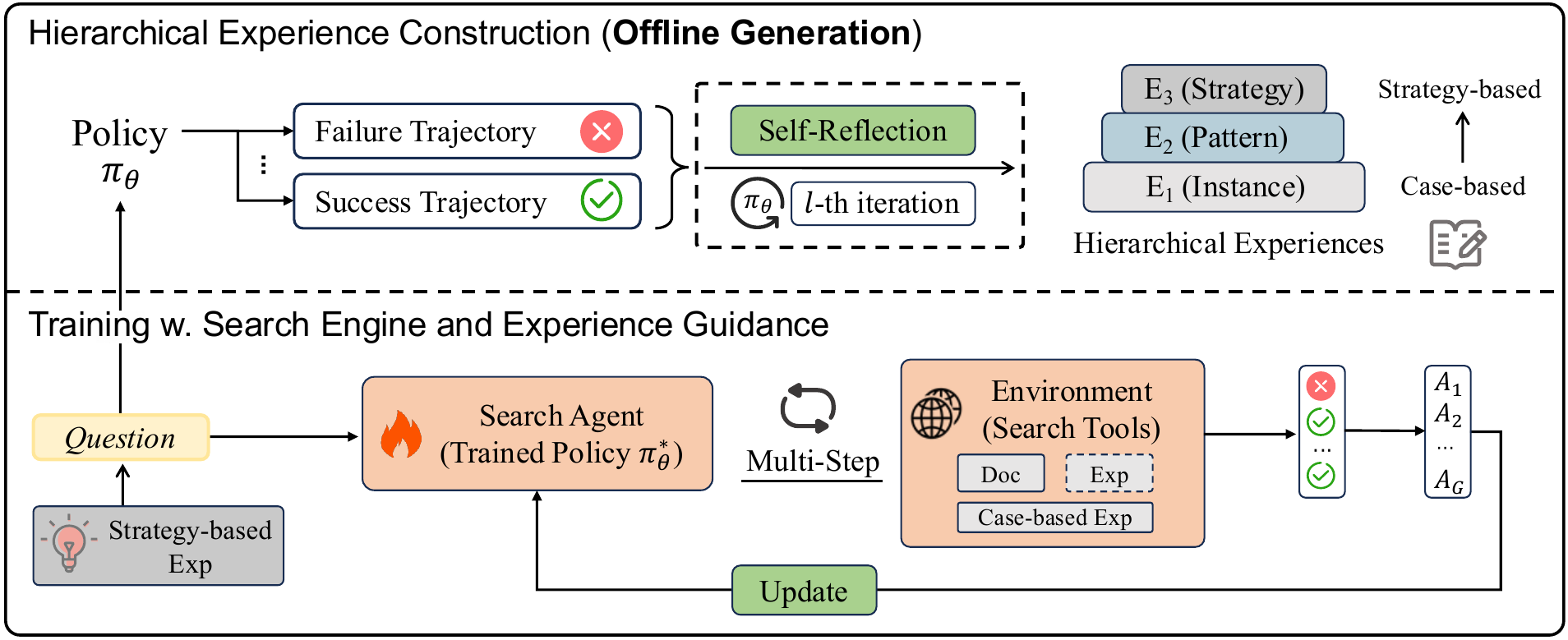} 
\caption{Overview of the offline hierarchical experience construction and the experience-guided policy optimization framework. The hierarchy spans from atomic instances to strategic principles, providing multi-granularity guidance for the search agent. During the training process, strategy-based experiences are leveraged to guide initial planning, while case-based experiences are employed to provide fine-grained support for intermediate reasoning steps.}
\label{framework}
\end{figure*}

\section{Methodology}
In this section, we propose a framework designed to transition the stochastic exploration inherent in critic-free RL algorithms toward experience-aligned heuristic search. Beyond leveraging external factual knowledge bases, we conceptualize the historical trajectories generated during the rollout phase as an endogenous knowledge base. As shown in Figure~\ref{framework}, the framework consists of two primary components:
\begin{itemize}
\item Hierarchical Experience Construction: This phase extracts success-critical features from raw trajectories through contrastive sampling and subsequently refines fragmented insights into systematic principles using clustering algorithms.

\item Experience-Aligned Training: This phase dynamically injects the distilled hierarchical knowledge into the training process of critic-free algorithms, effectively lifting the upper bound of the model's reasoning efficiency.
\end{itemize}

\subsection{Hierarchical Experience Construction} \label{sec:HEC}
In contrast to traditional static external knowledge sources, our HiExp framework introduces a self-evolving mechanism termed Self-Reflection Experience. This mechanism empowers the LLM to autonomously extract, abstract, and refine knowledge from its internal reasoning trajectories, as formalized in the hierarchical mining process of Algorithm~\ref{alg:HEC}. We further broaden the value of the training data beyond the annotated labels to encompass the entire exploration process.

\subsubsection{Contrastive Distillation}

For each sample $x_i$ in the training set, we execute $K$ independent rollouts to obtain a trajectory set $\mathcal{Y}_i$. Each trajectory comprises complete reasoning steps $\texttt{<think>}$, search actions $\texttt{<tool\_call>}$, and the corresponding external environment responses $\texttt{<tool\_response>}$. Guided by the outcome reward $r_{\mathrm{orm}}$, we partition these trajectories into successful paths $\mathcal{Y}_i^+$ and failed paths $\mathcal{Y}_i^-$. We leverage the self-reflection capabilities of either the policy model or a superior teacher model to identify two critical features: key decision points and reasoning traps. 
The output of contrastive distillation is formalized as case-based experience $e_i$ and its corresponding summary description $d_i$, which together encapsulate high-value procedural knowledge extracted from pre-sampled trajectories. Concretely, in the JSON output presented in Table~\ref{exp_contrastive_distillation}, the `description' field corresponds to $e_i$ and captures detailed experiential knowledge, whereas the `title' field corresponds to $d_i$ and functions as the retrieval key. In this way, raw rollout data are transformed into the foundational primitives for the subsequent hierarchical clustering phase.
\begin{align}
\mathrm{e}_i, \mathrm{d}_i = \text{Reflect}(x_i, y_i^+, y_i^-).
\end{align}

\subsubsection{Hierarchical Clustering}

Although case-based experiences $e_i$ extracted from contrastive trajectories encapsulate valuable reasoning clues, their instance-specific nature often limits direct utility. Direct injection into the LLM may trigger overfitting or introduce significant retrieval noise due to an expansive search space. To mitigate these challenges, we propose a multi-level clustering mechanism that transforms fragmented experiences into strategic experience knowledge.

First, we employ a pre-trained semantic encoder $\phi(\cdot)$ to map all case-based experiences into a high-dimensional embedding space. For each experience entry $e_i$, its vector representation is denoted as $\mathbf{v}_i = \phi(\mathrm{d}_i)$. This transformation ensures that semantically equivalent but lexically distinct experiences (e.g., verifying director identity vs. confirming director uniqueness) are proximal within the vector space. 
Subsequently, we apply agglomerative clustering to $\mathbf{v}_i$ for subsequent aggregation. The detailed procedure is provided in phase 2 of Algorithm~\ref{alg:HEC}. By imposing a stringent similarity threshold $\tau_1$, we consolidate experiences related to analogous problems. For each identified cluster $\mathcal{C}_j^{(1)}$, we leverage an LLM to distill multiple instance-level experiences into a generalized strategic experience. This distillation process is executed iteratively, leveraging systematic clustering and agentic self-reflection to progressively enhance the compactness and generalization of the experience repository.

\subsection{Experience-Aligned Training} \label{sec:EAT}
Inspired by Search-o1~\cite{li2025searcho1}, current advanced RAG methods introduce an agentic search strategy, transforming the exploration process into an iterative interaction between the intrinsic reasoning of LLMs and the external environment, thus effectively activating their autonomous reasoning capabilities. During interactions with the external environment, these methods often rely on unstructured text retrieval systems to supplement information for intermediate reasoning steps. Irrelevant textual noise can easily result in inefficient intermediate queries and logical drift. 
RL-based search agents~\cite{jin2025searchr1,chen2025research} typically rely on prior-free stochastic exploration during the rollout phase, which often suffers from low sample efficiency and limited convergence stability. 

To alleviate these bottlenecks and surpass the performance ceilings of vanilla exploration, we introduce an Experience-Aligned Guidance mechanism. This framework empowers the agent to dynamically leverage high-fidelity strategic priors from Hierarchical Experience Knowledge (HEK) during trajectory generation, effectively transforming undirected search into an experience-guided exploration process.
Within critic-free RL algorithms such as GRPO~\cite{shao2024deepseekmath}, the reasoning trajectory or intermediate query $q_t$ generated at each rollout step $t$ serves as a representation of the current state. The system utilizes a semantic encoder $\phi(\cdot)$ to compute the embedding vector of $q_t$, which is subsequently subjected to similarity matching against the hierarchical experience indices within the HEK. The retrieved experience $\mathrm{e}^*$ is defined as:
\begin{align}
    \mathrm{e}^* = \argmax_{e \in {\text{HEK}}} \text{cos\_sim}(\phi(q_t), \phi(\mathrm{d})).
\end{align}
Across different stages of the rollout process, our framework employs a dynamic guidance strategy. In the initial reasoning stage, global strategic experiences ($\mathrm{E}_2$ or $\mathrm{E}_3$) are prioritized and incorporated into the system prompt to provide strategic guidance that transcends specific task contexts, described in Table~\ref{system_template}. During intermediate reasoning steps, the system adaptively provides top-$\mathrm{k}$ granular $\mathrm{E}_1$ heuristics, filtered by a fixed semantic threshold to ensure high proximity to the current query.
The agent can also adaptively refine its sub-query planning. This hierarchical retrieval allows the model to leverage the self-reflection experience to effectively navigate complex multi-step reasoning paths, transforming stochastic exploration into an experience-aligned heuristic search.

Combining the outcome reward with GRPO training objective, we propose a RL objective that explicitly incorporates a search engine $\mathcal{R}$ and a hierarchical experience knowledge base ${\text{HEK}} = \{\mathrm{E}_{1}, \mathrm{E}_{2}, \dots, \mathrm{E}_{L}\}$ during optimization for the search agent training~\cite{jin2025searchr1}. Since reasoning trajectories are conditioned on hierarchical experiences, the advantage function derived from sampling trajectories possesses superior quality, facilitating more stable policy updates. The optimization objective is defined as:
\begin{multline}
    \mathcal{J}(\theta) =\mathbb{E}_{x \sim \mathcal{D}, \{y_i\} \sim \pi_{\text{old}}} \bigg[ \frac{1}{G} \sum_{i=1}^{G} \min \bigg(
    r_i(\theta) \hat{A}_i, \\
    \text{clip} \big( 
    r_i(\theta),\, 1-\epsilon,\, 1+\epsilon \big) \hat{A}_i \bigg) \bigg]  - \beta \mathbb{D}_{\text{KL}},
\end{multline}
where $r_i(\theta)=\frac{\pi_{\theta}(y_i|x,\mathrm{E}_l;\mathcal{R})}{\pi_{\text{old}}(y_i|x, \mathrm{E}_l;\mathcal{R})}$, $\pi_{\theta}$ denotes the trainable policy model, $\hat{A}_i$ represents the overall advantage function, $\mathbb{D}_{\text{KL}}$ denotes the KL divergence between the trained policy $\pi_{\theta}$ and the
reference policy $\pi_{\text{ref}}$ and $\beta$ is a hyper-parameter. $x$ are sampled from the dataset $\mathcal{D}$, and $y$ denote the output sequence interleaving reasoning steps with search engine retrievals. During the loss calculation phase, we mask all retrieved document snippets and case-based experiences within the intermediate reasoning steps to prevent the training policy from being biased.

\begin{table*}[ht]
\centering
\resizebox{1.0\textwidth}{!}{
\begin{tabular}{lccccccccccccl}
\toprule
\multirow{2}{*}{Methods} & \multicolumn{3}{c}{\textbf{HotpotQA$^\dagger$}}                               & \multicolumn{3}{c}{\textbf{2Wiki$^\dagger$}}  & \multicolumn{3}{c}{\textbf{Musique$^\dagger$}}   & \multicolumn{3}{c}{\textbf{Bamboogle$^\ddagger$}} & \multicolumn{1}{l}{\textbf{Average}}            \\ 
\cmidrule(r){2-4} \cmidrule(r){5-7} \cmidrule(r){8-10} \cmidrule(r){11-13} \cmidrule(r){14-14}
&F1 & CEM  & EM   & F1  & CEM  & EM & F1& CEM & EM & F1&CEM & EM &CEM\\ 
\hline
\rowcolor[rgb]{0.9,0.9,0.9}
\multicolumn{14}{c}
{\textbf{\textit{Prompt Based}}}  \\
\hline
\multicolumn{14}{l}
{\textbf{\textit{Qwen2.5-7B}}}    \\
Vanilla RAG                 & 29.0    & 22.4        & 20.5     & 32.5     & 27.9& 27.0 & 11.2 &5.1 &3.4 & 17.6 & 12.8 & 10.4  & 17.1     \\
$\text{Iter-RetGen}$      &   51.4    & 45.2    & 39.9      & 39.2     & 35.5       & 32.2  &17.4 &12.4&10.0&31.8&24.8&22.4&29.5      \\
$\text{IRCoT}$        &   47.2     & 47.3   & 35.3         & 35.0     & 39.2   & 25.5  &14.7 &13.3 &7.5 &32.3 &28.8 &23.2  &32.2       \\
$\text{Search-o1}^{*}$    & 44.4  & 41.2   &   34.2    & 50.8  & 51.0  & 41.8  & 18.1 &  15.5& 11.1 &  37.5 &31.2 &  27.2 & 34.7   \\
 \rowcolor[rgb]{0.8745, 0.9176, 0.9608}
\multicolumn{1}{l}
{{+ HiExp}}      & \textbf{48.7}& \textbf{47.7} & \textbf{37.1} & \textbf{54.8} &\textbf{56.8}  &\textbf{45.2} & \textbf{22.4}& \textbf{18.6} & \textbf{15.1} & \textbf{44.6} & \textbf{37.6}  & \textbf{33.6}  &\textbf{40.2} \scalebox{0.85}{$ (\uparrow\!5.5)$} \\ 
\midrule


\multicolumn{14}{l}
{\textbf{\textit{Frontier LLMs}}}    \\

$\text{DeepSeek-R1}$   & 62.5   & 54.0   & 48.0          & 65.7       & 65.0      & 54.0 &39.9&33.0&27.5&  63.0&52.8&52.0&51.2   \\

$\text{Qwen3-235B-A22B}$        & 57.3       & 56.1       & 44.5       & 59.4        & 64.1         & 45.3 &41.7&39.5&27.6&55.3&49.2&43.8&52.2          \\

$\text{GPT-4.1}$        & 60.6     & 56.0    & 45.0         & 69.7          & 75.5         & 56.0 &44.9 &47.0 &28.5 &63.8 &55.2 & 49.6&58.4       \\
$\text{o4-mini}$        & 57.8   & 59.5    &    40.5      & 62.1  &   71.0    & 47.5 & 41.6 &45.5 &27.5  &61.7 &64.0 &46.4 &60.0       \\
$\text{Gemini-2.5-Pro}$        & 55.6&60.5&39.5&71.8&83.0&60.5&37.0&47.0&24.5& 59.7& 69.6& 52.0&65.0     \\

\hline
\rowcolor[rgb]{0.9,0.9,0.9}
\multicolumn{14}{c}
{\textbf{\textit{Training Based}}}  \\
\hline
\multicolumn{14}{l}
{\textbf{\textit{Qwen2.5-7B}}}  \\
$\text{Search-R1-v0.3}$     & 61.8 &  53.6  & 49.8      & 60.7      & 58.7     & 52.3   & 30.9 &   24.7  & 21.5   & {59.4}&   48.0  & \textbf{47.2}  & 46.3   \\
$\text{ReSearch}$     & 63.2 &  55.8  & 50.4      & 67.1      & 65.4     & 60.3   & 28.0 &   34.1  & 24.0   & 53.1 &   45.6  & 41.6  & 48.7   \\
$\text{R1-Searcher}$     & 57.8 &  59.7  & 45.6     & 64.0     & 67.8    & 56.2   & 28.4 &  27.9  & 19.5   & 49.8 &   46.4  & 36.0  & 50.5   \\
\rowcolor[rgb]{0.8745, 0.9176, 0.9608}
\multicolumn{1}{l}
{{HiExp-Searcher}}   & \textbf{65.4} & \textbf{60.4} & \textbf{52.4} & \textbf{74.6} &\textbf{76.5}  &\textbf{66.9} & \textbf{41.7}& \textbf{36.7} & \textbf{30.7} & \textbf{61.0}& \textbf{50.4}  & \underline{46.4}  &\textbf{56.0} \scalebox{0.85}{$ (\uparrow\!9.7)$} \\

\midrule
\multicolumn{14}{l}
{\textbf{\textit{Qwen2.5-32B}}}  \\
$\text{Search-R1-v0.3}$  &66.5&55.8&53.5&73.4&71.7&68.1&36.2&30.6&28.5&65.1&55.2&54.4&53.5     \\
$\text{ReSearch}$  & 69.4 &61.0&56.3&78.1&76.7&72.3&39.3&33.8&30.5&63.1&52.0&50.4&55.9     \\

\rowcolor[rgb]{0.8745, 0.9176, 0.9608}
\multicolumn{1}{l}
{HiExp-Searcher}        & \textbf{71.2} & \textbf{62.9} & \textbf{57.8} & \textbf{81.5} & \textbf{80.4} & \textbf{75.8} & \textbf{49.2}& \textbf{41.1} & \textbf{36.2} & \textbf{68.2} & \textbf{57.2} & \textbf{54.8} & \textbf{60.4} \scalebox{0.85}{$ (\uparrow\!6.9)$} \\ 

\bottomrule
\end{tabular}

}
\caption{Overall evaluation results on the dev or test sets of four benchmarks. The best and second best results are bold and underlined, respectively. All methods are evaluated in the same local retrieval environment. * indicates the results reproduced by us. $^\dagger/\ddagger$ represents in-domain/out-of-domain datasets. $\mathtt{+}$ indicates architectural updates, such as base model replacement or new module integration.
}
\label{primary_table}
\end{table*}

\begin{table}[ht!]
\centering
\setlength{\tabcolsep}{0.31mm}
\resizebox{0.47\textwidth}{!}{
\begin{tabular}{l *{4}{>{\centering\arraybackslash}p{30pt}}}
\toprule
\multirow{2}{*}{Methods} & \multicolumn{2}{c}{\textbf{In-Domain}}  &   \multicolumn{2}{c}{\textbf{Out-of-Domain}}                           
\\ \cmidrule(r){2-3} \cmidrule(r){4-5} &  F1   & CEM     & F1    & CEM         \\ \midrule
\textbf{Training-free} \\
(a) Doc Search only  & 37.8 & 35.9  & 31.3 &  24.2  \\
(b) w/ $\mathrm{E}_2$+$\mathrm{E}_1$ & \textbf{42.0} & \textbf{41.0} & \textbf{34.8} & \textbf{27.3}  \\
(c) w/ $\mathrm{E}_3$+$\mathrm{E}_1$ & 41.0 & 38.9 & 33.4 &  26.8 \\
\midrule

\textbf{Training} \\
(a) Baseline GRPO     & 54.2 & 49.6   & 34.4 &  30.3 \\
(b) w/ $\mathrm{E}_2$      & 59.3  & 56.8 & 36.8 &  33.6     \\
(c) w/ $\mathrm{E}_3$      &  56.1 & 53.2  & 34.8 & 31.5      \\
(d) w/ $\mathrm{E}_2$+$\mathrm{E}_1$ & \textbf{60.6}&\textbf{ 57.9 } & \textbf{38.2} & \textbf{34.0} \\
(e) w/ $\mathrm{E}_3$+$\mathrm{E}_1$ & 57.7 & 53.8 & 35.5 &  32.7     \\
\bottomrule 
\end{tabular}
}
\caption{Ablation study on various multi-hop datasets. "w/" represent "with". Performance evaluations for all trained models utilize the corresponding optimal retrieval configurations of HEK and document.}
\label{ablation}

\end{table}

\begin{table}[ht!]
\centering
\setlength{\tabcolsep}{1.21mm}
\resizebox{0.47\textwidth}{!}{
\begin{tabular}{lccccccc}
\toprule
\multirow{2}{*}{Methods} & \multicolumn{2}{c}{\textbf{Bam$^\ddagger$}}  &   \multicolumn{2}{c}{\textbf{Frames$^\ddagger$}}  &   \multicolumn{2}{c}{\textbf{MoreHQA$^\ddagger$}}   &\multicolumn{1}{c}{\textbf{Avg.}}                             
\\ \cmidrule(r){2-3} \cmidrule(r){4-5} \cmidrule(r){6-7} \cmidrule(r){8-8} 
        & F1    & LasJ     & F1    & LasJ     & F1    & LasJ  & LasJ \\ \midrule
Search-o1        & 60.4 &  66.2  & 26.6 & 33.5   & 25.4 & 36.6  &   45.4   \\
ReSearch      & 71.9  & 73.8 & 38.7 & 48.5 & 30.9 & 46.5  &56.3  \\ 
R1-Searcher   & 67.2  & 71.3 & 33.4 & 42.6 & 23.5 & 37.9 &    50.6  \\ \midrule
\textbf{Ours}            & \textbf{75.6} & \textbf{76.4}&\textbf{41.3} &  \textbf{48.7} & \textbf{34.7} &  \textbf{49.4} & \textbf{58.2} \\ \bottomrule
\end{tabular}
}
\caption{Generalization experiments on out-of-domain datasets using online search engine.}
\label{web_search_table}
\end{table}

\section{Experiments}
\subsection{Experimental Settings}
\noindent\textbf{Datasets and Evaluation Metrics}. We conduct extensive experiments on six multi-hop datasets, including HotpotQA~\cite{yang2018hotpotqa}, 2WikiMultiHopQA (2Wiki)~\cite{ho20202wiki}, Musique~\cite{trivedi2022musique}, Bamboogle (Bam)~\cite{press2022bam}, MoreHopQA (MoreHQA)~\cite{schnitzler2024morehopqa}, and Frames~\cite{krishna2024frames}. The first three datasets are in-domain datasets, with portions of their training sets used for training, while the latter three are out-of-domain datasets utilized to evaluate the model's generalization performance. Our evaluation is conducted on the full dev or test sets corresponding to the above datasets. For evaluation metrics, we employ the standard word-level F1 score (F1), Cover Exact Match (CEM), and Exact Match (EM). For more complex open-domain QA tasks, we additionally utilize LLM-as-Judge (LasJ) to ensure a fair evaluation. 
To evaluate domain generalization, we also perform experiments on six mathematical reasoning benchmarks, including AIME 2024/2025, AMC~\cite{li2024amc}, MATH-500~\cite{hendrycks2021math500}, Minerva~\cite{lewkowycz2022Minerva}, and OlympiadBench~\cite{he2024olympiadbench}. For AIME, AMC benchmarks with a limited number of samples, we report $\text{Avg}@32$ over 32 independent runs; for others, we use $\text{Pass}@1$ metric.

\begin{table*}[ht]
\centering
\setlength{\tabcolsep}{1.51mm}
\begin{tabular}{lccccccl}
\toprule
\multirow{1}{*}{Methods} & \multicolumn{1}{c}{\textbf{AIME24}} & \multicolumn{1}{c}{\textbf{AIME25}}  &   \multicolumn{1}{c}{\textbf{AMC}}  &   \multicolumn{1}{c}{\textbf{MATH500}} &   \multicolumn{1}{c}{\textbf{Minerva}}   &   \multicolumn{1}{c}{\textbf{Olympia}}  &\multicolumn{1}{l}{\textbf{Average}} \\ \midrule                     
 $\text{Base Model}$       & 13.2 &  6.1  & 44.5 & 58.4  &25.4 & 29.0  &  29.4   \\
+ HiExp    &  12.8 & 7.3 & 42.7 & 62.6 & 25.7  & 32.0 & 30.5 \scalebox{0.85}{$ (\uparrow\!1.1)$}\\ 
\midrule
SFT     & 21.7  & 16.7 & 55.4 & 82.8 & 36.4 & 45.3 & 43.1 \scalebox{0.85}{$ (\uparrow\!13.7)$} \\  
GRPO     &  24.1 & 17.8 & 58.8 & 83.4 & 35.3 & 47.4  & 44.5 \scalebox{0.85}{$ (\uparrow\!15.1)$}  \\
GRPO + HiExp  &26.7   & 23.3 & 62.7 & 84.2 & 37.1 & 46.8  &\textbf{46.8}  \scalebox{0.85}{$ (\uparrow\!17.4)$}\\ 
\bottomrule
\end{tabular}
\caption{Performance comparison across six mathematical reasoning benchmarks on Qwen2.5-Math-7B. "+ HiExp" uses the proposed hierarchical experience at inference phase without any training. "GRPO + HiExp" incorporates HiExp during GRPO training.}
\label{comp_math}
\end{table*}

\noindent\textbf{Search Tools}.
An efficient search tool is essential for our search agent. We build a local retrieval environment using a dense retriever with the multilingual-e5-base~\cite{wang2022e5} model, incorporating the 2018 Wikipedia corpus~\cite{ho20202wiki}. To obtain more up-to-date information, we further utilize Tavily as a web search tool.

\noindent\textbf{Baselines and Training Details}.
In our experiments, in addition to comparing with state-of-the-art LLMs such as \textit{DeepSeek-R1-0528}, \textit{Qwen3-235B-A22B}, \textit{GPT-4.1-0414}, o4-mini-0416, and \textit{Gemini-2.5-Pro-0325} (as shown in Table~\ref{primary_table}), we also benchmark against advanced RAG methods~\cite{shao2023iterretgen,trivedi2022ircot,li2025searcho1} and RL-based agentic search models~\cite{jin2025searchr1,chen2025research,song2025r1searcher,zheng2025deepresearcher,wang2025stepsearch}. These experiments are primarily based on the Qwen2.5 models~\cite{qwen2025qwen25technicalreport}, where Qwen2.5-7B and Qwen2.5-32B refer to their respective Instruct models. All training-based models are derived from their corresponding open-source models. 

The training data of search agent consist of the stage-2 data from \citet{song2025r1searcher} and 8,000 randomly sampled instances from Musique. For mathematical reasoning tasks, we  train on the OpenR1-Math 45k subset~\cite{openr1,yan2025luffy}. We utilize FSDP~\cite{zhao2023pytorchfsdpexperiencesscaling} and vLLM~\cite{vllm} in VeRL~\cite{verl} framework, with a sampling temperature of 1.0, top-p of 0.95 and a maximum response length of 8192. The detailed training process are shown in Appendix~\ref{details}.

\subsection{Main Results}
Table~\ref{primary_table} provides a comprehensive evaluation of HiExp-Searcher against several strong baselines on four multi-hop benchmarks, while demonstrating the performance gains achieved by our approach when integrated into a prompt-based paradigm.

\noindent\textbf{Achieves continuous performance gains}. 
Our approach achieves significant performance improvements on multiple complex multi-hop benchmarks under all evaluation metrics. Unlike previous RL-based search agents that struggle with inefficient reasoning trajectories or redundant computations, HiExp-Searcher effectively guides the reasoning path to achieve a superior balance between response comprehensiveness and accuracy. Furthermore, our method is designed as a universal and pluggable enhancement that can be seamlessly integrated into various agentic frameworks and retrieval environments to achieve further performance boosts.

\noindent\textbf{Achieve frontier LLM performance with small-scale models}. We evaluate current state-of-the-art LLMs on several multi-hop reasoning benchmarks. Interestingly, we observe that these frontier models do not consistently benefit from the Search-o1 series prompts for multi-step reasoning and retrieval. Therefore, for these large models, we adopt a standard RAG setup to obtain stronger and more stable performance. A key contribution of our framework is the ability to empower small-scale models (e.g., 7B or 32B) to match or exceed the reasoning capabilities of much larger, frontier LLMs. Our trained 7B model achieves performance on par with GPT-4.1 and surpasses larger LLMs like DeepSeek-R1 and Qwen3-235B-A22B. These results demonstrate that our method can effectively bridge the capability gap between compact models and frontier systems.

\begin{table}[ht]
\centering
\setlength{\tabcolsep}{1.21mm}
\resizebox{0.48\textwidth}{!}{
\begin{tabular}{lccccccl}
\toprule
\multirow{2}{*}{Methods} & \multicolumn{2}{c}{\textbf{HotpotQA}}  &   \multicolumn{2}{c}{\textbf{2Wiki}}  &   \multicolumn{2}{c}{\textbf{Musique}}   &\multicolumn{1}{l}{\textbf{Average}}                             
\\ \cmidrule(r){2-3} \cmidrule(r){4-5} \cmidrule(r){6-7} \cmidrule(r){8-8} 
        & F1    & CEM     & F1    & CEM     & F1    & CEM  & CEM \\ \hline
 $\text{Search-o1}^{*}$   & 44.4&  41.2  & 50.8 & 51.0    & 18.1  &   15.5 &   35.9    \\
\ \ + HiExp      &  48.7 &  47.7  & 54.8 &  56.8  & 22.4 &  18.6  &  \textbf{41.0}  \scalebox{0.85}{$(\uparrow\!5.1)$}  \\
\midrule
GRPO      & 61.6  & 54.9 & 63.6 & 61.2 & 37.4 & 32.8  &  49.6\\ 
\ \ + HiExp    &  65.4 & 60.4 &74.6  & 76.5 & 41.7  &  36.7 & \textbf{57.9}\scalebox{0.85}{$(\uparrow\!8.3)$} \\ 
GSPO    & 52.3 & 62.7 & 57.9 & 60.0 & 29.6 & 35.7 &  52.8  \\
\ \ + HiExp & 56.9 & 64.7 & 62.8 & 69.4&36.7 & 42.6 & \textbf{58.9} \scalebox{0.85}{$(\uparrow\!6.1)$}  \\ 
\bottomrule
\end{tabular}
}
\caption{Performance comparison of HiExp integrated with different RL algorithms.}
\label{comp_rl_algorithms}
\end{table}

\subsection{Further Analysis}
\subsubsection{Ablation Studies}
The ablation study results presented in Table~\ref{ablation} underscore the substantial performance gains achieved by integrating HEK into training-free and training-based settings. In the training-free category, the inclusion of strategy-based ($\mathrm{E}_2$) and case-based ($\mathrm{E}_1$) experiences significantly elevates the in-domain F1 score from 37.8 to 42.0 and the CEM from 35.9 to 41.0, demonstrating the plug-and-play capability of the HiExp framework. This trend is even more pronounced in the training phase, where the full HEK configuration ($\mathrm{E}_2$+$\mathrm{E}_1$) propels the baseline GRPO's performance from an in-domain F1 of 54.2 to a peak of 60.6, with a corresponding CEM increase from 49.6 to 57.9. These improvements validate that experience-aligned optimization effectively internalizes complex reasoning logic, allowing the agent to transcend the limitations of stochastic exploration and coarse outcome rewards.

A comparative analysis of experience granularities reveals that pattern-level induction ($\mathrm{E}_2$) provides more effective guidance than higher-level $\mathrm{E}_3$, particularly when combined with instance-level corrections ($\mathrm{E}_1$). Across all benchmarks, the configuration $\mathrm{E}_2$+$\mathrm{E}_1$  consistently outperforms the alternative $\mathrm{E}_3$+$\mathrm{E}_1$ in the training section, achieving an out-of-domain F1 of 38.2 and a CEM of 34.0, which notably exceeds the out-of-domain GRPO baseline results of 34.4 and 30.3 respectively. This disparity suggests that specific task-structure patterns are more actionable for the model during the reasoning process than abstract meta-rules. Furthermore, the robust gains on out-of-domain datasets confirm that the framework facilitates the acquisition of generalized reasoning blueprints rather than mere memorization, ensuring high performance and stability even when encountering unfamiliar knowledge environments.

\begin{figure*}[t]
\centering
\includegraphics[width=1.00\textwidth]{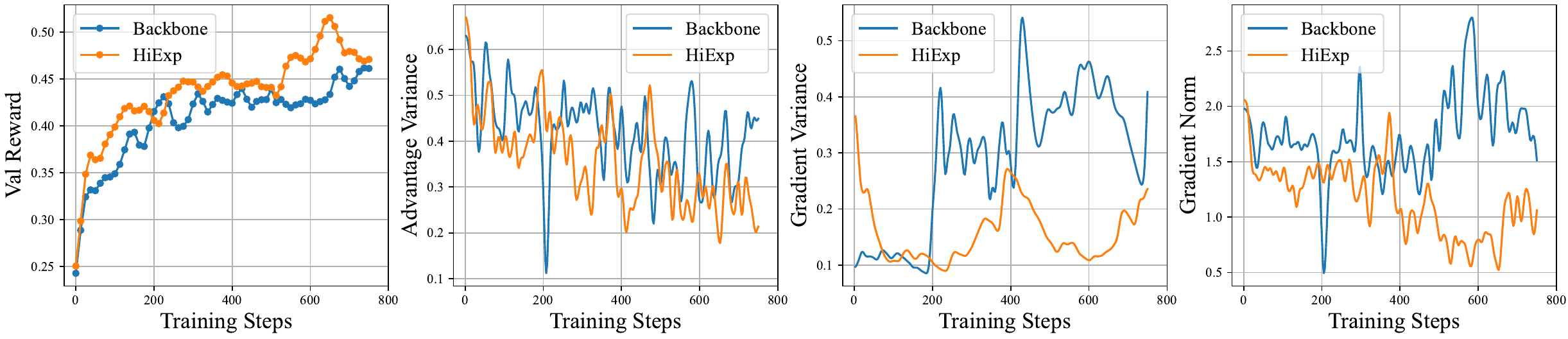} 
\caption{Training stability analysis of HiExp on multi-step retrieval benchmarks. Backbone denotes the performance of the base model trained via GRPO.}
\label{stability}
\end{figure*}

\subsubsection{Generalization Performance Analysis}
\noindent\textbf{Cross-Task and Out-of-Domain Generalization}. The framework demonstrates significant versatility by extending beyond multi-hop question answering into mathematical reasoning tasks. Experimental results in Table~\ref{comp_math} show that integrating HiExp during GRPO training yields a substantial gain of +17.4 over the base model, while achieving consistent performance gains in out-of-domain scenarios. This advancement highlights that the hierarchical distillation of experience, specifically the transition from raw trajectories to strategic meta-principles, reinforces the model's fundamental logical processing and enables it to excel in domains requiring rigorous reasoning. Even in training-free scenarios, the addition of HiExp provides a consistent performance boost, confirming that the acquired experience-guided paradigms are robust enough to improve the model's reasoning ceiling without requiring further parameter updates.

\noindent\textbf{Cross-Algorithm Generalization}. The pluggable nature of the HiExp allows for significant performance gains across multiple RL algorithms. When integrated with various algorithms such as GRPO and GSPO~\cite{zheng2025gspo}, the framework yields consistent gains in CEM scores. This broad applicability across different optimization strategies and tasks confirms that experience-guided alignment is a generalizable solution for maximizing the training upper bound and inference reliability of agentic search agents.

\noindent\textbf{Cross-Environment Generalization}. The external information retrieval environment serves as a fundamental component for search agents. We also evaluate the agent in more realistic interactions by incorporating web search as shown in Table~\ref{web_search_table}. The experimental results indicate that web search provides substantial performance gains by offering diverse and dynamic context. Throughout the training phase, the experience-guided mechanism within HiExp-Searcher effectively steers the agentic search process through hierarchical planning and grounding.

\subsubsection{Experience Source Analysis}
To examine whether our framework benefits more from stronger external teachers or from self-generated experiences, we compare self-distillation with strong-teacher distillation in Table~\ref{tab:teacher_self_distill}. In the self-distillation setting, the base policy model (Qwen2.5-7B) generates its own experiences for subsequent training. In the strong-teacher setting, we replace these experiences with those generated by a larger model, Qwen-Max, while keeping the downstream hierarchical organization and RL training pipeline unchanged.
The results show that self-distillation slightly outperforms strong-teacher distillation in multi-hop reasoning tasks, with an average improvement of about 1.2\%. This finding suggests that, for our framework, the effectiveness of the method is driven less by the absolute quality of the initial reflections and more by the compatibility between the generated experiences and the student model’s reasoning distribution. The experiences produced by the 7B model itself appear to be better aligned with its capability boundary, making them easier to interpret and exploit during RL training. This observation also supports the practical value of self-distillation, which enables a fully self-contained and scalable training pipeline without relying on external large-model supervision.

\begin{table}[t]
\centering
\small
\begin{tabular}{lcccc}
\toprule
\textbf{Source} & \textbf{Hotpot} & \textbf{2Wiki} & \textbf{Musique} &\textbf{Avg.} \\
\midrule 
Max $\rightarrow$ 7B & 59.8 & 74.7 & 35.5 & 56.7 \\
7B $\rightarrow$ 7B (Self) & \textbf{60.4} & \textbf{76.5} & \textbf{36.7} & \textbf{57.9}  \\
\midrule
\textbf{Gain} & \textbf{+0.6} & \textbf{+1.8} & \textbf{+1.2} & \textbf{+1.2} \\
\bottomrule
\end{tabular}
\caption{Ablation on experience source. Self-distillation slightly outperforms strong-teacher distillation.}
\label{tab:teacher_self_distill}
\end{table}

\subsection{Training Stability Analysis}\label{sec:stability}
To evaluate the training stability of our framework, we analyze the evolution of reward signals and the variance across group rollouts, demonstrating the definitive advantages of hierarchical experience knowledge over the stochastic exploration typically observed in Figure~\ref{stability}. 
The proposed HiExp framework facilitates a more rapid and stable ascent in valid reward by leveraging hierarchical experience guidance to steer the policy model toward high-value reasoning paths. Unlike the stochastic exploration inherent in traditional reinforcement learning, which often leads to factually inconsistent queries and inefficient trajectories, HiExp ensures that sampled rollouts remain aligned with distilled reasoning principles. This strategic alignment effectively avoids the noisy or redundant trajectories that typically plague agentic search systems.

Consequently, our approach significantly reduces the variance in both advantages and gradients compared to the baseline. By providing more consistent and stable advantage estimates ($\hat{A}_i$) throughout the optimization process, HiExp suppresses gradient noise and stabilizes model updates. This improved stability allows the model to internalize efficient search behaviors more effectively, ultimately pushing the performance ceiling higher across diverse reasoning tasks.

\subsection{Qualitative Analysis}
To gain a deeper understanding of how hierarchical experience knowledge transforms the LLM's internal reasoning, we conduct a qualitative analysis of
our experience-guided agent in Table~\ref{case_study}. 
In this case, the strategy-based experience $\mathrm{E}_2$ provides the "logic blueprint" by identifying a multi-hop constraint decomposition strategy. Instead of searching for plays from May 2016 in isolation, the experience instructs the model to resolve the temporal anchor first: identifying Natalie Diaz as the author of "Postcolonial Love Poem" (winning the MacArthur Fellowship in 2018), thereby establishing 2018 as the target fellowship year for the playwright. Simultaneously, the case-based $\mathrm{E}_1$ experience serves as a "surgical correction" to maintain search precision during the execution phase. For example, once the agent identifies Dominique Morisseau as a 2018 MacArthur Fellow, $\mathrm{E}_1$ prevents the common trap of confusing a play's premiere date with its specific composition or publication month, ensuring the model accurately targets the work written in May 2016.
From an optimization perspective, this qualitative precision directly translates into the training stability discussed in Section~\ref{sec:stability}. By suppressing redundant steps, HiExp provides more consistent and stable advantage estimates throughout the RL training process.

\section{Conclusions}
In this paper, we propose HiExp, an endogenous hierarchical experience construction framework tailored for search agents, which synthesizes meta-knowledge through self-reflection and agglomerative clustering over internal reasoning trajectories. HiExp facilitates the autonomous distillation of experiential priors, ensuring logical consistency while eliminating external data dependencies. Our framework not only bolsters LLM performance across diverse tasks during the inference phase but also dynamically aligns with the rollout stage of reinforcement learning. This alignment effectively transforms conventional stochastic exploration into a strategic, experience-guided search, significantly enhancing the stability and effectiveness of policy optimization. Extensive evaluations demonstrate that HiExp consistently yields substantial performance gains over state-of-the-art RL-based agents, exhibiting robust generalization across diverse task domains and reinforcement learning algorithms.

\section*{Limitations} 
Despite the substantial improvements in reasoning accuracy and training stability, our current framework possesses certain limitations that offer promising avenues for future research. Our current approach operates in a semi-decoupled manner, where the construction of hierarchical experience is isolated from the subsequent policy optimization. This static approach implies that the guidance distilled from the initial policy model may fail to synchronize with the model's evolving capabilities as training progresses. As the agent internalizes more sophisticated reasoning paradigms through reinforcement learning, it may encounter higher-order challenges.
Therefore, a crucial future direction lies in establishing a dynamic closed-loop system where experience construction and model training are tightly coupled.

\bibliography{custom}

\clearpage
\appendix
\section{Implementation Details}\label{details}
\begin{figure*}[t]
\centering
\includegraphics[width=1.0\textwidth]{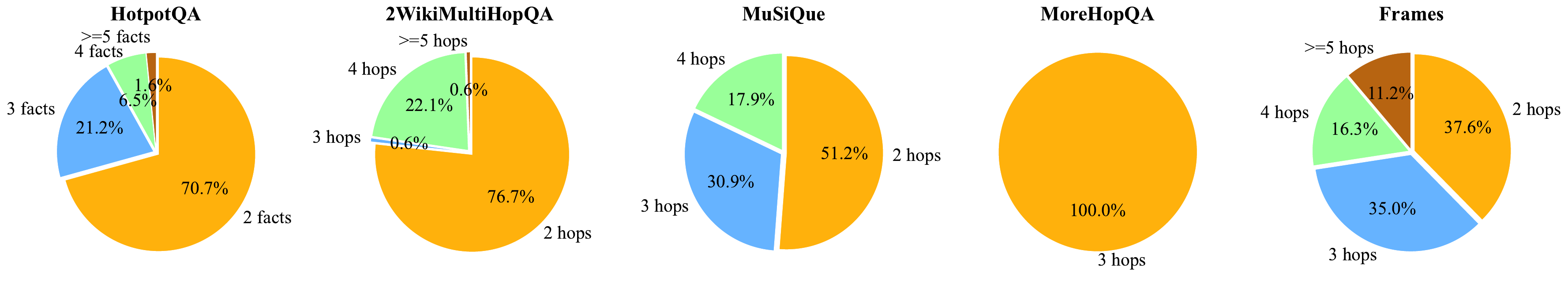} 
\caption{Overview of the distribution of query complexity over five multi-hop QA datasets.}
\label{hop_dist}
\end{figure*}

Due to the large size of the dev sets in the 2WikiMultiHopQA and HotpotQA datasets, which affects iteration efficiency, we randomly sample 1,000 examples from their respective dev sets as our final test set, with a fixed random seed 42. We also verify that the performance on this subset is nearly identical to that on the full dev set, indicating that this approach can significantly improve iteration efficiency. To
better understand the complexity of multi-hop reasoning in these datasets, we analyze the hop distribution of the HotpotQA, 2WikiMultiHopQA, MuSiQue, MoreHopQA, and Frames dev/test sets in Figure~\ref{hop_dist}. The statistics show that there is a high proportion of complex reasoning queries with 3 hops or more. HotpotQA lacks explicit hop annotations, so we instead count the number of supporting facts. 
During the hierarchical experience construction, we employ the trained policy model for contrastive distillation and subsequent clustering of hierarchical experiences, thereby avoiding the introduction of external supervisory signals and enabling self-driven capability iteration and knowledge distillation.

In the retrieval process, we employ \text{multilingual-e5-base} as the retriever and use the widely used Wikipedia
dump from December 2018 as the retrieval corpus, which comprises over 21 million passages. To improve retrieval efficiency, we combine the supporting document passages from five multi-hop datasets with one million randomly sampled documents from the 2018 Wikipedia dump to create our final retrieval corpus. 
All HEK are encoded using the same embedding model. We adopt a parent-child retrieval architecture, where succinct summary descriptions serve as child chunks for semantic matching. Upon a successful match, the corresponding detailed experiences are retrieved as parent chunks to provide the necessary context for the reasoning process. We apply a 0.8 similarity threshold for case-based experiences ($\mathrm{E}_1$) to ensure high precision. For strategy-based experiences ($\mathrm{E}_2$ or $\mathrm{E}_3$), we select the top-5 candidates to maintain a diverse set of guidance strategies.

During the training phase of search agent, our training data consist of a total of 8,148 examples from HotpotQA and 2WikiMultiHopQA, which are selected through data selection in R1-Searcher. In addition, we randomly sample 8,000 examples from the training set of MuSiQue to form our final training set. The training consists of 2 epochs, with a \verb|train_batch_size| of 16 and a learning rate of 1e-6. \verb|ppo_mini_batch_size| is set to 16. The maximum lengths for prompt and response are set to 512 and 8192. Rollouts are conducted with a batch size of 8 and a temperature of 1.0 to encourage exploration. The KL-divergence regularization coefficient and the clipping ratio are set to 1e-3 and 0.2, respectively. All experiments are carried out on eight NVIDIA-H20-96G. In the inference stage, we use SGLang or vLLM as the underlying inference engines and set different maximum context lengths and maximum retrieval times to avoid the impact of outlier samples on training. For the evaluation of other prompt-based baselines, we use the implementations provided in the ReSearch GitHub repository\footnote{\url{https://github.com/Agent-RL/ReCall}.}. For other training-based methods, we evaluate them using their publicly available trained models.

\section{Prompt Examples}
Table~\ref{judge_prompt} presents the implementation prompt for our LLM-as-Judge (LasJ) score. By leveraging larger and more powerful LLMs as judge models, we can achieve more accurate judgments of responses in the multi-hop QA scenarios. This more precise evaluation approach can be incorporated into the training process, which also introduces additional computational overhead for training. The prompts used for contrastive distillation and subsequent clustering of hierarchical experiences are provided in Tables~\ref{exp_contrastive_distillation} and \ref{exp_cluster}.

\section{Quantitative Analysis}
Table~\ref{primary_table} in the main content presents the performance of state-of-the-art LLMs on multi-hop question answering tasks. Interestingly, we find that these models struggle to effectively follow instructions under the search-o1 paradigm, resulting in suboptimal performance. Additionally, in the basic RAG setting, where models are simply asked to answer questions based on retrieved documents, the models tend to respond that no relevant information is found when the answer is not present in the retrieved documents, failing to fully utilize their inherent capabilities. Therefore, after optimizing the prompt for the RAG scenario (see Table~\ref{rag_plus_prompt}), the models are able to better integrate their own fundamental abilities with the retrieved information to jointly solve the original questions.

We also provide a detailed analysis of the computational overhead introduced by the offline experience construction pipeline in Table~\ref{tab:offline_cost}. This pipeline consists of two main stages: (1) contrastive distillation over pre-sampled trajectories, and (2) hierarchical clustering for organizing the distilled experiences. On the MusiQue dataset, using approximately 7,000 initial trajectories, the contrastive distillation stage requires about 1 hour with Qwen-7B/72B/Max under parallel inference. The subsequent hierarchical clustering stage is lightweight, taking around 10 minutes on CPU using \texttt{scikit-learn}'s agglomerative clustering with Ward linkage. In total, the complete offline pipeline incurs less than 2 GPU-hours of equivalent cost. Compared with the subsequent RL optimization stage, this overhead is relatively small. In our setting, GRPO training requires approximately 36 GPU-hours (using 8$\times$H20 GPUs for 1 epoch). Therefore, the offline experience construction phase accounts for less than 6\% of the total computation budget.

\begin{algorithm}[t!]
\caption{Hierarchical Experience Construction}
\label{alg:HEC}
\begin{algorithmic}[1]
\Require Training set $\mathcal{D}$, rollout count $K$, reward function $R$, max depth $L_{max}$
\Ensure Hierarchical Experience Knowledge base ${\text{HEK}} = \{\mathrm{E}_{1}, \mathrm{E}_{2}, \dots, \mathrm{E}_{L}\}$

\State Initialize atomic experience set $\mathrm{E}_{1} \gets \emptyset$

\Statex \textbf{// Phase 1: Contrastive Experience Extraction}
\For{each sample $x_i \in \mathcal{D}$}
    \State $\mathcal{Y}_i \gets \text{Sample\_K\_Rollouts}(x_i, K)$
    \State $\mathcal{Y}_{pos}, \mathcal{Y}_{neg} \gets \text{Split\_By\_Reward}(\mathcal{Y}_i, R)$ \Comment{Contrastive splitting}
    \If{$\mathcal{Y}_{pos} \neq \emptyset$ \textbf{and} $\mathcal{Y}_{neg} \neq \emptyset$}
        \State $\omega \gets \text{LLM\_Contrast}(x_i, \mathcal{Y}_{pos}, \mathcal{Y}_{neg})$ \Comment{Extract success-critical insights}
        \State $\mathrm{E}_{1} \gets \mathrm{E}_{1} \cup \{\omega\}$
    \EndIf
\EndFor
\Statex \textbf{// Phase 2: Self-Reflection \& Iterative Hierarchical Abstraction}
\State ${\text{HEK}} \gets \{\mathrm{E}_{1}\}$
\For{$l = 2$ \textbf{to} $L_{max}$}
    \State $Z \gets \text{Encoder}(\mathrm{E}_{l-1})$ \Comment{Project insights from previous level}
    \State $\mathcal{C}_{local}$ $\gets$ $\text{Agglomerative\_Clustering}$$(Z,$\allowbreak$ \text{threshold}$$=\tau_l)$ \Comment{Semantic clustering}

    
    \State $\mathrm{E}_{l} \gets \emptyset$
    \For{each cluster $c \in \mathcal{C}_{local}$}
        \State $\phi $$\gets$$\text{LLM\_Summarize\_}$$\text{Cluster}$$(c, \text{level}$\allowbreak$=l)$ \Comment{Pattern induction for current level}
        \State $\mathrm{E}_{l} \gets \mathrm{E}_{l} \cup \{\phi\}$
    \EndFor
    
    \State ${\text{HEK}} \gets {\text{HEK}} \cup \{\mathrm{E}_{l}\}$ \Comment{Append new level to knowledge base}
    
    \If{$|\mathrm{E}_{l}| \le 1$}
        \State \textbf{break} \Comment{Convergence: Global principles reached}
    \EndIf
\EndFor

\State \Return ${\text{HEK}}$
\end{algorithmic}
\end{algorithm}

\begin{table}[ht!]
\centering
    \renewcommand{\arraystretch}{0.8}
\small
\begin{tabular}{p{7.2cm}}
\toprule
\toprule
In this environment you have access to a set of tools you can use to assist with the user query. \
You may perform multiple rounds of function calls. \
In each round, you can call one or more functions. \

Here are available functions in JSONSchema format:
```json\verb|\n|\{\text{func\_schemas}\}\verb|\n|'''
\newline

Here are some relevant reasoning experience and examples to guide your decision-making process: 

\{\text{experience}\}
\newline

In your response, you need to first think about the reasoning process in the mind and then conduct function calling to get the information or perform the actions if needed. \
The reasoning process and function calling are enclosed within \texttt{<think>} \texttt{</think>} and \texttt{<tool\_call>} \texttt{</tool\_call>} tags. \
The results of the function calls will be given back to you after execution, \
and you can continue to call functions until you get the final answer for the user's question. \
Finally, if you have got the answer, enclose it within \verb|\boxed{}| with latex format and do not continue to call functions, \
i.e., \texttt{<think>} Based on the response from the function call, I get the weather information. \texttt{</think>} The weather in Beijing on 2025-04-01 is \verb|\boxed{20C}|.
\newline

For each function call, return a json object with function name and arguments within \texttt{<tool\_call>}\texttt{</tool\_call>} XML tags:

\texttt{<tool\_call>}\verb|\n|\{\{"name": $\text{<function-name>}$, "arguments": $\text{<args-json-object>}$\}\}\verb|\n|\texttt{</tool\_call>}

\\
\bottomrule
\bottomrule
\end{tabular}
\caption{System prompt for generating reasoning trajectories through interaction with the environments during training and inference stages.}
\label{system_template}
\end{table}

\begin{table}[h]
\centering
\small
\begin{tabular}{ll}
\toprule
\textbf{Component} & \textbf{Cost} \\
\midrule
Contrastive Distillation & $\sim$1 hour \\
Hierarchical Clustering & $\sim$10 min (CPU) \\
Offline Pipeline Total & $< 2$ GPU-hours eq. \\
GRPO Training & $\sim$36 GPU-hours \\
Offline Cost Ratio & $< 6\%$ \\
\bottomrule
\end{tabular}
\caption{Computational cost of the offline experience construction pipeline on MusiQue.}
\label{tab:offline_cost}
\end{table}

\begin{table}[t!]
\centering
    \renewcommand{\arraystretch}{0.8}
\small
\begin{tabular}{p{7.2cm}}
\toprule
\toprule

  \textbf{Question}: What is the name of the play written in May 2016 by a playwright who won the MacArthur Fellowship the same year as the poet who wrote ``Postcolonial Love Poem''? 
 \hfill \textbf{Answer}: \text{Skeleton Crew} \\
\midrule
\texttt{\textbf{HiExp}}: \textbf{\texttt{<think>}} [{Trigger $\mathrm{E}_2$ Experience: Multi-hop Constraint Decomposition}] This query requires a strict sequence: identify the poet $\rightarrow$ determine their award year $\rightarrow$ find a playwright from the same cohort $\rightarrow$ verify a specific play from May 2016. \textbf{\texttt{</think>}}

\textbf{\texttt{<tool\_call>}} \{``query'': ``which poet wrote the book `Postcolonial Love Poem'?''\} \textbf{\texttt{</tool\_call>}}

\textbf{\texttt{<tool\_response>}} 
... Natalie Diaz is the author of Postcolonial Love Poem. In 2018, she received a MacArthur Fellowship ... \textbf{\texttt{</tool\_response>}}\\
\textbf{\texttt{<think>}} ... \textbf{\texttt{</think>}}\\

\textbf{\texttt{<tool\_call>}} \{``query'': ``who is the playwright that won the MacArthur Fellowship in 2018?''\} \textbf{\texttt{</tool\_call>}}\\

\textbf{\texttt{<tool\_response>}} [Trigger $\mathrm{E}_1$ Experience: Temporal-Professional Attribute Intersection] ... \textbf{\texttt{</tool\_response>}}\\
\textbf{\texttt{<think>}} ... \textbf{\texttt{</think>}}\\

\textbf{\texttt{<tool\_call>}} \{``query'': ``what is the name of the play written by Dominique Morisseau in May 2016?''\} \textbf{\texttt{</tool\_call>}}\\

\textbf{\texttt{<tool\_response>}} [{Trigger $\mathrm{E}_1$ Experience: Temporal-Entity Work Attribution}] ... \textbf{\texttt{</tool\_response>}}\\
\textbf{\texttt{<think>}} ... \textbf{\texttt{</think>}}\\

\textbf{\texttt{<answer>}} The final answer is \text{\textbackslash boxed\{Skeleton Crew\}} \textbf{\texttt{</answer>}}\\

\bottomrule
\bottomrule
\end{tabular}
\caption{Quantitative analysis of the efficient reasoning process in Frames dataset.}
\label{case_study}
\end{table}

\begin{table}[ht]
\centering
\renewcommand{\arraystretch}{0.8}
\small
\begin{tabular}{p{7.2cm}}
\toprule
\toprule
You are an expert in question answering. Given a question within \texttt{<question>} \texttt{</question>} \
and some contexts within \texttt{<context>} \texttt{</context>}, you first think about the reasoning process within \texttt{<think>} \texttt{</think>} \
and put the answer within \texttt{<answer>} \texttt{</answer>}. \\
For example, \texttt{<question>} This is a question \texttt{<question>} \
\texttt{<context>} Here are contexts \texttt{<context>} \texttt{<think>} This is the reasoning process. \texttt{</think>} \
\texttt{<answer>} The final answer is \text{\textbackslash boxed\{ answer here \}}\texttt{</answer>}. \
If the answer could not be deduced from the contexts or it's wrong, give the right answer based on your own knowledge. \
If the question is ambiguous or the contexts contain multiple possible answers, \
list all possible answers within \verb|\boxed{}| with latex format, separated by commas.
\\
\bottomrule
\bottomrule
\end{tabular}
\caption{Prompt for vanilla retrieval augmented generation.}
\label{rag_plus_prompt}
\end{table}

\begin{table}[ht]
\centering
    \renewcommand{\arraystretch}{0.8}
\small
\begin{tabular}{p{7.2cm}}
\toprule
\toprule
An agent system is provided with a set of experiences and has tried to solve the question multiple times with both successful and wrong solutions. Review these problem-solving attempt and extract generalizable experiences.
Follow these steps:
\newline

1. Trajectory Analysis:

- For successful steps: Identify key correct decisions, insights and formats used

- For errors: Pinpoint where and why the reasoning, answer or formatting went wrong

- Note any important patterns or strategies used\/missed 

- Review why some trajectories fail? Is there any key steps are missed, or formats are wrong?

2. Experiences Summarization:

- Summarize and output with the following format:

\{

  \ \ \ \ "type": "The category to classify the question, including domain and solving method",
  
  \ \ \ \ "title": "A one-sentence summary of the general experience",
  
  \ \ \ \ "tags": ["Key words or tags, fewer than 5 words"],
  
  \ \ \ \ "description": "Your analysis here, within 100 words",
  
  \ \ \ \ "thinking": "Your thinking process here, especially comparing correct and incorrect solution attempts, within 100 words"
  
\}
\\
\bottomrule
\bottomrule
\end{tabular}
\caption{Prompt for contrastive distillation.}
\label{exp_contrastive_distillation}
\end{table}

\begin{table}[ht]
\centering
    \renewcommand{\arraystretch}{0.8}
\small
\begin{tabular}{p{7.2cm}}
\toprule
\toprule
You are given a set of experiences that an agent has accumulated while solving various questions.  
Your task is to cluster the similar experiences into generalized experiences that capture the core patterns and strategies.  
These generalized experiences should enable the agent to solve similar questions correctly and efficiently in the future.

The set of experiences is listing with the following format:
[\{"type": "", "title": "", "tags": "", "description": "", "thinking": "", "qa\_groups": [{"id": "", ...}]\}, ...], where qa\_groups is the questions and answers in this experience group.
\newline

Summarize and output with the following format:

[\{
  "ids": ["all qa ids from the experiences in this cluster"],
  "type": "A category for this group of questions, including domain and solving method.",
  "title": "A one-sentence summary of the generalized strategy for this cluster.",
  "tags": ["A list of up to 5 keywords or tags."],
  "description": "Your analysis of the common patterns and core logic for this cluster, within 100 words.",
  "thinking": "Your thinking process here, especially differences within the group of experiences, within 100 words"
\}]
\\
\bottomrule
\bottomrule
\end{tabular}
\caption{Prompt for hierarchical experience clustering.}
\label{exp_cluster}
\end{table}

\clearpage
\begin{table}[ht!]
\centering
    \renewcommand{\arraystretch}{0.8}
\small
\begin{tabular}{p{7.2cm}}
\toprule
\toprule
You will be provided with three pieces of content: the questioner's question, the user's response, and the reference answer list.
Your task is to score the accuracy of the user's response based on the criteria outlined below.
Please ensure that you carefully read and understand these instructions.\\
Evaluation Criteria:\\
Accuracy - Whether the user's answer is consistent with the reference answer and answers the questioner's question. We define this dimension as "whether the user's response includes all the key points from the reference answer and answers the questioner's question."\\
Evaluation Steps:\\
1. Carefully read the questioner's question and understand its key points.\\
2. Carefully read the reference answer and understand the key points relevant to the question.\\
3. Check whether the user's response includes all the key points from the reference answer and answers the questioner's question.\\
4. Based on the evaluation criteria, assign a score in the range of 0 to 5, where 0 indicates that the user's response does not include any of the key points from the reference answer and completely fails to answer the questioner's question; 5 indicates that the user's response includes all the key points from the reference answer and fully and correctly answers the questioner's question.\\
Example:\\
Questioner's question:\\
\{question\}\\
Reference answer:\\
\{answer\}\\
User's response:\\
\{response\}

Evaluation result (output only the score between 0 and 5):
\\
\bottomrule
\bottomrule
\end{tabular}
\caption{Judge prompt for LLM-as-judge scoring.}
\label{judge_prompt}
\end{table}

\end{document}